\documentclass[10pt,twocolumn,letterpaper]{article}

\usepackage[pagenumbers]{cvpr} 

\usepackage{algorithmic}
\usepackage{algorithm}
\usepackage{textcomp}
\usepackage{stfloats}
\usepackage{url}
\usepackage{verbatim}
\usepackage{booktabs}       
\usepackage{multirow}
\usepackage{pifont}         
\usepackage{xcolor}         
\usepackage[normalem]{ulem}
\useunder{\uline}{\ul}{}
\usepackage{times}
\usepackage{epsfig}
\usepackage{amsmath}
\usepackage{amssymb}
\usepackage{array}
\usepackage{float}
\usepackage{wrapfig}
\usepackage{caption}
\usepackage{enumitem}
\usepackage[normalem]{ulem}
\useunder{\uline}{\ul}{}
\usepackage{graphicx}
\usepackage{subcaption}
\usepackage{adjustbox}
\usepackage[table]{xcolor}
\definecolor{mygray}{rgb}{0.9, 0.95, 1}

\definecolor{cvprblue}{rgb}{0.21,0.49,0.74}
\usepackage[pagebackref,breaklinks,colorlinks,allcolors=cvprblue]{hyperref}

\title{Revealing the Implicit Noise-based Imprint of Generative Models}

\author{
Xinghan Li\\
Fudan University\\
Shanghai, China\\
{\tt\small xinghanli24@m.fudan.edu.cn}
\and
Yue Yu\\
Fudan University\\
Shanghai, China\\
{\tt\small yueyu24@m.fudan.edu.cn}
\and
Xue Song\\
Fudan University\\
Shanghai, China\\
{\tt\small xuesong21@m.fudan.edu.cn}
\and
Haijun Shan\\
CEC GienTech Technology Co.,Ltd.\\
Shanghai, China\\
{\tt\small haijun.shan@gientech.com}
\and
Jingjing Chen\thanks{Corresponding author.}\\
Fudan University\\
Shanghai, China\\
{\tt\small chenjingjing@fudan.edu.cn}
}

\begin{document}
\maketitle
\begin{abstract}
With the rapid advancement of vision generation models, the potential security risks stemming from synthetic visual content have garnered increasing attention, posing significant challenges for AI-generated image detection. 
Existing methods suffer from inadequate generalization capabilities, resulting in unsatisfactory performance on emerging generative models. To address this issue, this paper presents \textbf{NIRNet} (\textbf{N}oise-based \textbf{I}mprint \textbf{R}evealing \textbf{Net}work), a novel framework that leverages noise-based imprint for the detection task. 
Specifically, we propose a novel Noise-based Imprint Simulator to capture intrinsic patterns imprinted in images generated by different models. By aggregating imprint from various generative models, imprint of future models can be extrapolated to expand training data, thereby enhancing generalization and robustness. Furthermore, we design a new pipeline that pioneers the use of noise patterns, derived from a Noise-based Imprint Extractor, alongside other visual features for AI-generated image detection, significantly improving detection performance. Our approach achieves state-of-the-art performance across seven diverse benchmarks, including five public datasets and two newly proposed generalization tests, demonstrating its superior generalization and effectiveness.
\end{abstract}
\section{Introduction}
\label{sec:intro}

The advancement of deep generative models has led to a remarkable improvement in synthesized image quality. Notably, recent methods~\cite{GAN,CycleGAN,BigGAN,LDM,DiT,DPM} have demonstrated impressive capabilities in generating photorealistic images that are nearly indistinguishable from real ones. While these developments have achieved both academic and commercial success, they have also given rise to serious risks, such as misinformation propagation, identity fraud and copyright violations~\cite{xu2023combating, yan2025gpt, yan2024df40,barrett2023identifying}. Therefore, it is crucial to develop robust and generalizable detectors to distinguish AI-generated images.

\begin{figure}[t]
    \centering
    \includegraphics[width=0.9\columnwidth]{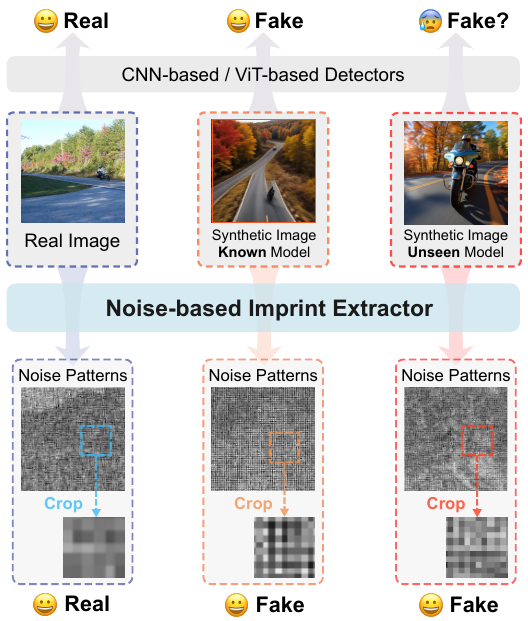}
    \caption{Conventional detectors (Top) perform well on known generative models, but typically struggle with images from unseen architectures. Our approach (Bottom) utilizes a \textbf{Noise-based Imprint Extractor} to capture a universal generative \textbf{noise patterns}, distinct from the real one, ensuring robust generalization.}
    \label{fig:noise_pattern}
\vspace{-2mm}
\end{figure}

Existing detection approaches~\cite{Fusing,LGrad,DIRE,NPR,UnivFD,FatFormer, C2p-clip,zheng2024breaking} have demonstrated effectiveness when applied to known generative models. For example, works~\cite{UnivFD,AIDE} utilizing large pre-trained vision-language models like CLIP~\cite{CLIP} to capture semantic features show promising results. However, these methods exhibit significant performance deterioration when challenged with unseen synthesis technologies.
This limitation stems from training data bias, where detectors learn model-specific artifacts of known generators instead of universal generative patterns. Although some methods~\cite{rajan2024effectiveness,guillaro2024bias, chen2025dual} attempt to align data by injecting forensic artifacts into real images for training, the result remains heavily dependent on the architectures used, hindering their generalization to unseen models. Therefore, the critical challenge lies in developing detection methodologies that overcome these data-induced biases to achieve effective performance across diverse generative paradigms, particularly as new techniques continue to emerge.

To address the issues mentioned above, we propose a novel solution. Analogous to how camera sensor and optical process~\cite{lukas2006digital,zhong2023patchcraft} imprint unique physical noise patterns on real images, we posit that generative models~\cite{DDPM,LDM} inevitably introduce distinct noise patterns during the synthesis process, which differ from those in real images. As shown in Figure~\ref{fig:noise_pattern}, the noise patterns of real images and AI-generated images are inconsistent. We refer to it as ``\textbf{Noise-based Imprint}". These patterns are independent of the image semantic content and are considered an imprint left by the generative model during synthesis, serving as reliable cues for detection. Therefore, to enhance the generalization capability of the detector, we shift the focus from learning task-specific features to modeling fundamental noise discrepancies inherent in the synthesis process itself. 

\begin{figure}[t]
    \centering
    \includegraphics[width=0.75\columnwidth]{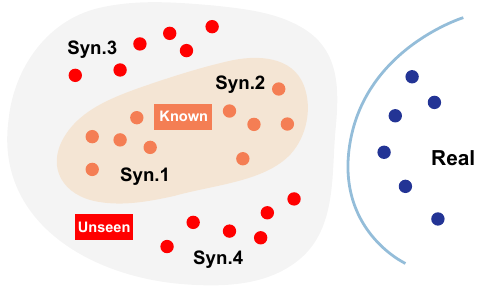}
    \caption{Conventional detectors tend to overfit to \textbf{Known} generative models (orange area), leading to poor performance on \textbf{Unseen} models (red dots). Our method, using a \textbf{Noise-based Imprint Simulator}, learns an extrapolated synthetic boundary (grey area) that covers unseen models, thus achieving robust generalization.}
    \label{fig:decision-boundary}
\vspace{-3mm}
\end{figure}

To this end, we propose \textbf{NIRNet} (\textbf{N}oise-based \textbf{I}mprint \textbf{R}evealing \textbf{Net}work), a novel and generalizable framework that leverages noise-based imprint for detecting AI-generated images. The framework comprises two consecutive stages: a simulation stage and a training stage. In the simulation stage, we first introduce a novel \textbf{Noise-based Imprint Simulator}, which reconstructs real images with various generative models. The difference between the original and reconstructed images is modeled as the \textit{noise-based imprint}, representing artificial traces left by the synthesis process. Subsequently, we sample from the fused distribution obtained by fusing the imprint distributions from different models. Through this approach, real images are transformed to incorporate noise-based imprint and subsequently used as negative samples in the training process, reducing data-induced biases to enhance generalization capabilities and robustness, as conceptually illustrated in Figure~\ref{fig:decision-boundary}.
In the training stage, we introduce a specialized \textbf{Noise-based Imprint Extractor} and design a novel detection pipeline. We propose a dedicated training strategy with an auxiliary objective to explicitly guide Noise-based Imprint Extractor to capture the imprint. Furthermore, a discriminator with hybrid features is employed, integrating noise feature maps from the extractor together with frequency and semantic features. This design enables precise discrimination between real and synthetic images, even for those generated by previously unseen models.

Overall, our contributions are summarized as follows:
\begin{itemize}
    \item We propose \textbf{NIRNet} (Noise-based Imprint Revealing Network), a novel framework including a Noise-based Imprint Simulator that captures intrinsic synthesis patterns across diverse generative models, enabling training data extrapolation and shifting the detection focus from model-specific artifacts to universal noise patterns.
    \item We propose a novel detection pipeline that integrates noise-based imprint with frequency and semantic features to achieve more discriminative representations.
    \item Extensive experiments demonstrate that NIRNet achieves state-of-the-art performance across five public benchmarks including GenImage~\cite{genimage}, Synthbuster~\cite{synthbuster}, Chameleon~\cite{AIDE}, SynthWildx~\cite{cozzolino2024raising} and WildRF~\cite{WildRF}. Moreover, we present two new evaluation datasets, Gen-8K and ForenGen, to assess detectors’ generalization capabilities under cross-model and in-the-wild scenarios.
\end{itemize}

\section{Related Work}
\label{sec:related_work}

With the rapid development of generative models, distinguishing between real and AI-generated images has become increasingly challenging. As a result, the demand for detecting AI-generated images is growing.

\textbf{Detection via Inherent Artifact Analysis.} Early approaches focused on handcrafted artifacts such as reflections~\cite{o2012exposing}, color~\cite{mccloskey2018detecting}, co-occurrence~\cite{nataraj2019detecting}, and saturation~\cite{mccloskey2019detecting}. However, these methods often struggle with generalization as generative models evolve. Later works trained CNN-based detectors (e.g., CNNSpot~\cite{CNNSpot}) to directly classify real and AI-generated images, showing limited but notable cross-generator generalization. Frequency-based approaches~\cite{FreDect,LNP,AIDE,zhong2023rich,FreqNet,Fire,SPAI,SAFE} identify significant artifacts in the frequency domain of synthesized images, which are caused by the up-sampling operations in the generation process. Gradient-based methods~\cite{LGrad} convert images into gradient-based representations, utilizing gradients as a generalized form of artifacts produced by generative models. Semantic-based methods~\cite{UnivFD,cozzolino2024raising} show that linear probing of a pre-trained, frozen CLIP image encoder can effectively detect fake images generated by a wide range of models. Methods such as LOTA~\cite{wang2025lota} further explore bit-plane decomposition to isolate subtle noisy representations. Additionally, reconstruction-based method~\cite{DIRE} employ DDIM inversion to reconstruct images and then train a classifier to detect differences. Another work~\cite{ricker2024aeroblade} extends to Latent Diffusion Models, using VAE-based reconstructions and LPIPS distance for detection.

\begin{figure*}[t]
\vspace{-1mm}
    \centering
    \includegraphics[width=\textwidth]{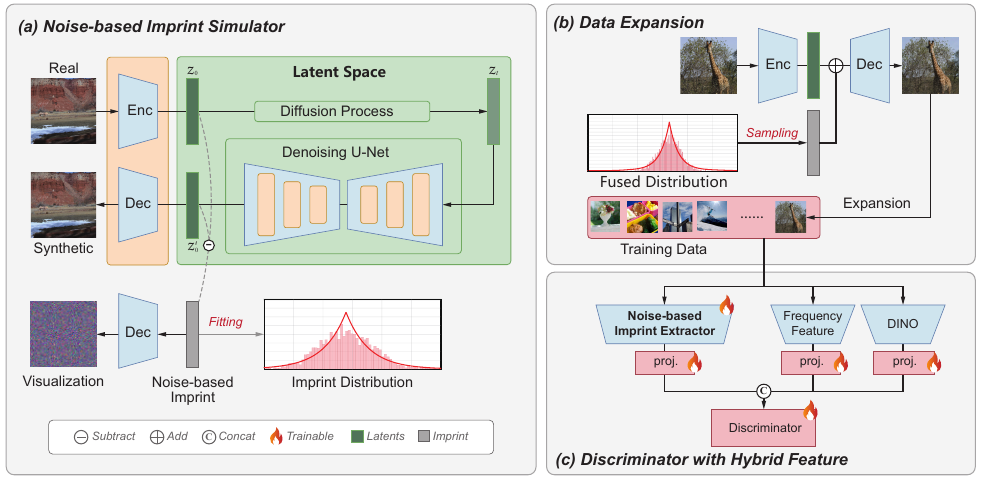}
    \caption{\textbf{Overall framework of the proposed NIRNet.} Our NIRNet consists of two stages: a simulation stage and a training stage. In the simulation stage (a) and (b), a Noise-based Imprint Simulator models the noise-based imprint of the generative model. This process computes differences after reconstruction, which are learned and fitted to a Laplace distribution. Subsequently, samples are drawn from a fused distribution derived from multiple models to transform real images into images embedded with imprint, thereby expanding the training dataset. In the training stage (c), end-to-end training is performed on the expanded dataset. A Noise-based Imprint Extractor, utilized to capture the intrinsic noise patterns of images, is introduced. In conjunction with frequency and semantic features, our framework functions in a hybrid feature manner to detect AI-generated images.}
    \label{fig:framework}
\vspace{-2mm}
\end{figure*}

\textbf{Detection via Simulated Artifact Injection.} Apart from the aforementioned methods, artifact injection-based methods are closely related to ours. Prior research has explored creating simulated fake images by injecting generative process traces into real images. For GAN-based detection, FingerprintNet~\cite{jeong2022fingerprintnet} reconstructs real images and injects synthesized frequency-domain fingerprints. In the context of diffusion-generated image detection, DRCT~\cite{chen2024drct} employed DDIM~\cite{DDIM} inversion to reconstruct both authentic and synthetic images, incorporating both types into contrastive learning frameworks during training. Study~\cite{rajan2024effectiveness} utilized the VAE~\cite{VAE} autoencoder from LDM to reconstruct real images without engaging the diffusion model's denoising procedure, thereby introducing artifacts inherent to the VAE. $D^3$~\cite{D3} injects discrepancy cues via a parallel distorted-feature branch, while Dual-Alignment~\cite{chen2025dual} aligns synthetic images with real ones across both pixel and frequency domains via VAE reconstruction. The approach in~\cite{guillaro2024bias} integrated autoencoder techniques with diffusion steps, leveraging the inpainting diffusion model of Stable Diffusion 2.1~\cite{Rombach2022stable} to generate self-synthesized images. In contrast, our approach models a fused artifact distribution from multiple generators. By sampling from this distribution to extrapolate training data, we enhance diversity and improve generalization to unseen models.
\section{Method}
\label{sec:method}

As illustrated in Figure~\ref{fig:framework}, our NIRNet comprises two stages: a \textbf{Noise-based Imprint Simulator} for data expansion and a \textbf{Noise-based Imprint Extractor} for end-to-end detection.

\subsection{Simulator and Data Expansion}\label{subsec:simulator}

\noindent \textbf{Noise-based Imprint Simulator.} 
We hypothesize that AI-generated images contain subtle noise patterns, or "imprint," left by the synthesis process. Our core idea is to model this noise-based imprint in the latent space, first for a single model, and then create a fused distribution to simulate imprints of unseen models, thereby enhancing generalization.

\subsubsection{Modeling the Imprint of a Single Generative Model}\label{subsec:model_imprint}
We obtain the imprint by reconstructing a real image $I$ and measuring the latent difference. We first encode $I$ to $\mathbf{z_0}$ via a VAE~\cite{VAE}, then reconstruct it using a diffusion model to get  $\mathbf{z_0'}$. The noise-based imprint is defined as the latent difference:
\begin{equation}
\Delta \mathbf{z} = \mathbf{z_0'} - \mathbf{z_0}.
\label{eq:delta_z}
\end{equation}

To build a statistical model of this imprint, we apply this to $n$ real images, collecting a set of latent difference tensors $\mathbf{X} \in \mathbb{R}^{n \times C \times H \times W}$, where $C$, $H$ and $W$ denote the number of channels, height and width, respectively. This $\mathbf{X}$ represents the characteristic imprint distribution in the latent space, rather than simple pixel-level reconstruction errors.

We model $\mathbf{X}$ with a Laplace distribution. We empirically found that a Laplace distribution fits the tails of the data better than a Gaussian (see Supplementary), which aligns with the sparse, heavy-tailed perturbations common in denoising processes. The mean $\boldsymbol{\mu} \in \mathbb{R}^{C \times H \times W}$ and standard deviation $\boldsymbol{\sigma} \in \mathbb{R}^{C \times H \times W}$ are computed along the dimension $n$:
\begin{equation}
    \boldsymbol{\mu}_{c,h,w} = \frac{1}{n}\sum_{i=1}^{n}\mathbf{X}_{i,c,h,w},
\label{eq:mean}
\end{equation}
\begin{equation}
    \boldsymbol{\sigma}_{c,h,w} = \sqrt{\frac{1}{n}\sum_{i=1}^{n}(\mathbf{X}_{i,c,h,w}-\boldsymbol{\mu}_{c,h,w})^2}.
\label{eq:sigma}
\end{equation}
The scale parameter $\mathbf{b}$ of the Laplace distribution is then computed as $\mathbf{b} = \frac{\boldsymbol{\sigma}}{\sqrt{2}}$. Thus, the imprint of a single model is modeled as:
\begin{equation}
    \mathbf{X} \sim \text{Laplace}(\boldsymbol{\mu}, \mathbf{b}).
\label{eq:laplace}
\end{equation}

While this approach effectively models the imprint of a single, known generator, our ultimate goal is to generalize to unseen models. This motivates the development of a fused imprint distribution.

\begin{figure}[t]
    \centering
    \begin{subfigure}[b]{0.49\columnwidth}
        \centering
        \includegraphics[width=\columnwidth]{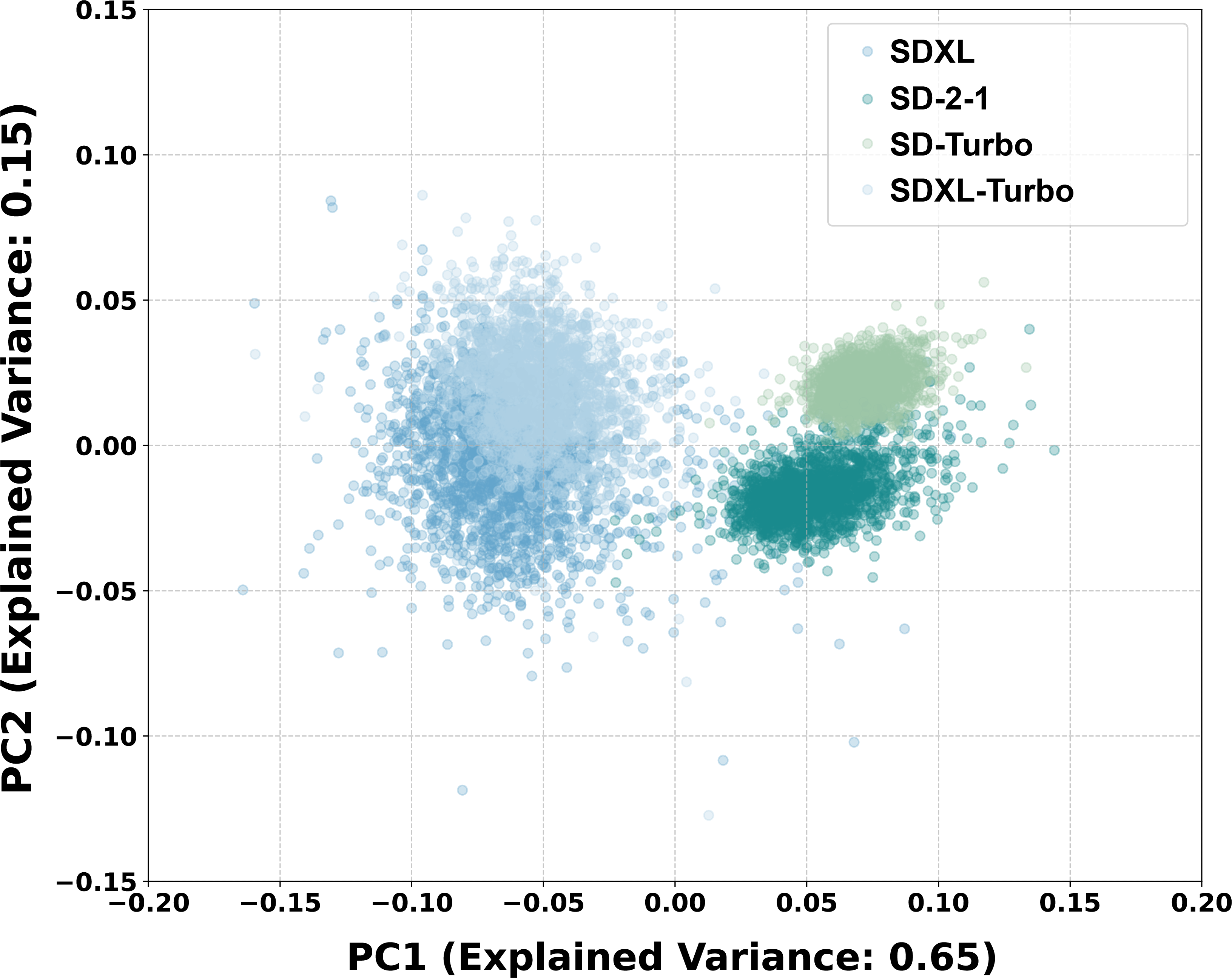}
        \caption{Distribution of noise-based imprint from four different models.}
        \label{fig:pca_base}
    \end{subfigure}
    \hfill
    \begin{subfigure}[b]{0.49\columnwidth}
        \centering
        \includegraphics[width=\columnwidth]{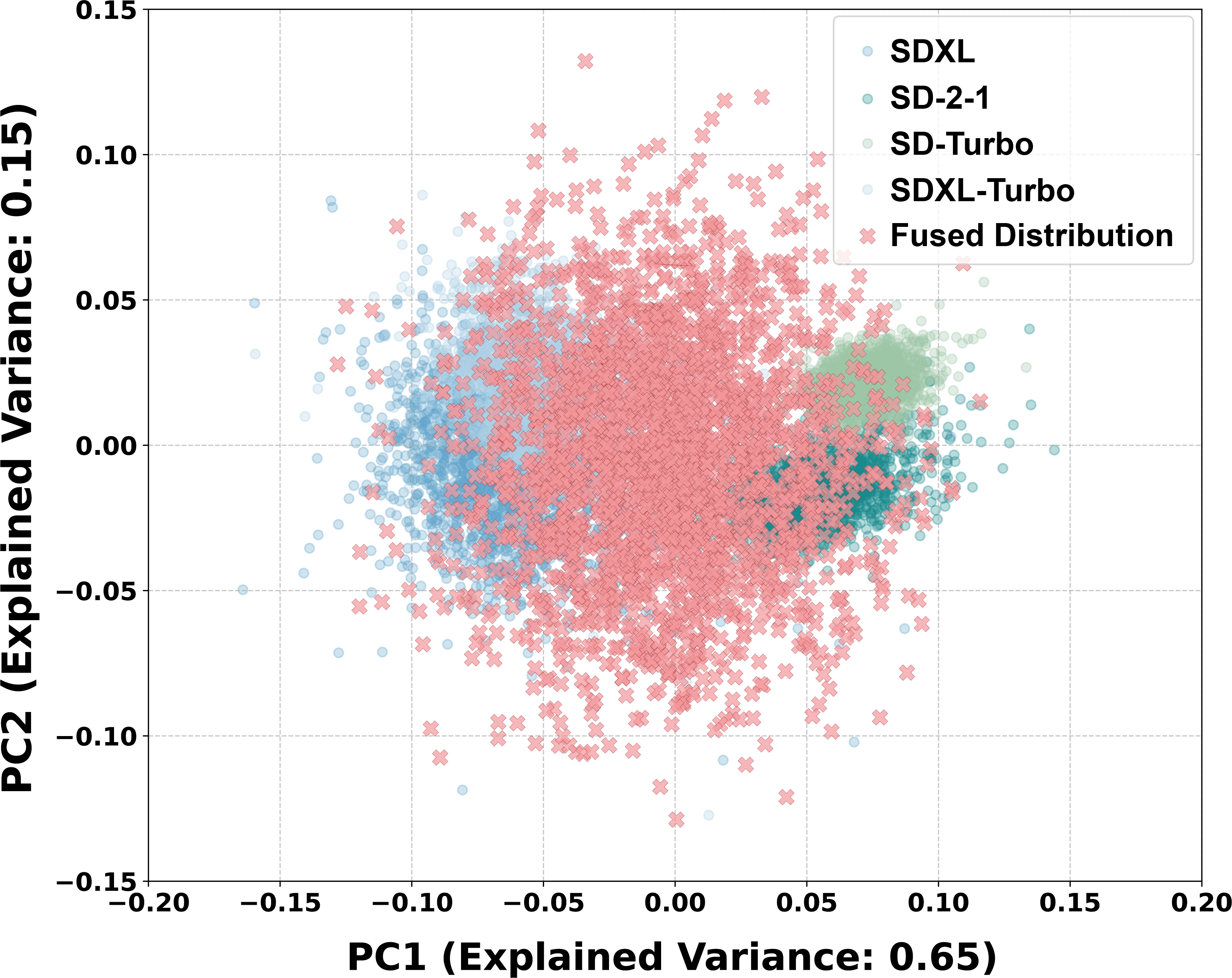}
        \caption{Visualization of the fused distribution.}
        \label{fig:pca_all}
    \end{subfigure}
    \caption{\textbf{PCA visualizations of noise-based imprint distributions} (a) from four different models, and (b) with the addition of a fused distribution used in Noise-based Imprint Simulator. The fused distribution helps simulate unseen or future model imprint.}
    \label{fig:pca_analysis}
\vspace{-2mm}
\end{figure}

\subsubsection{Simulating Unseen Models with a Fused Imprint Distribution}
To simulate imprints from unseen models, we construct a fused distribution from $m$ known generators. Our analysis (Figure~\ref{fig:pca_base}) reveals three key observations: (1) imprint distributions vary substantially across models rather than following a single unified distribution; (2) generators with similar architectures exhibit relatively minor differences in their distributions; and (3) these distributions are not entirely disjoint, showing considerable overlap. These observations motivate our approach: by fusing the individual distributions to form a composite distribution $X'$ (Figure~\ref{fig:pca_all}), we can span a broader, more generalizable imprint space. We hypothesize that this fused distribution can effectively simulate imprint from unseen or future generators.

Specifically, for each model $M_i$, we follow ~\cref{subsec:model_imprint} to obtain its latent difference tensor $\mathbf{X}^{(i)} \in \mathbb{R}^{n \times C \times H \times W}$. We compute a fused tensor by taking a weighted average:
\begin{equation}
    \mathbf{X}_{\text{fused}} = \sum_{i=1}^{m} w_i \cdot \mathbf{X}^{(i)},
\end{equation}
where $\sum_{i=1}^{m} w_i = 1$. We use uniform weights ($w_i = 1/m$). We then compute the mean $\boldsymbol{\mu}$ and scale $\mathbf{b}$ from $\mathbf{X}_{\text{fused}}$ (as in \cref{eq:mean,eq:sigma,eq:laplace}) to define our fused imprint distribution:
\begin{equation}
    \mathbf{X}' \sim \text{Laplace}(\boldsymbol{\mu}, \mathbf{b}). \label{eq:mixed_dist}
\end{equation}
This $\mathbf{X}'$ serves as our generalized imprint model, allowing us to simulate imprint from unseen generators.

\subsubsection{Expansion of Training Data}\label{subsec:expansion}
We leverage the fused imprint distribution $\mathbf{X}'$ for a data expansion strategy designed to extrapolate to unseen generators. The core idea is to synthesize novel training instances that exhibit imprints \textit{beyond} those of the specific generators in our known set.

In practice, for any given real image, we first encode it to obtain latent representation $\mathbf{z}_0$. We then sample an imprint perturbation $\Delta \mathbf{z^*}$ from $\mathbf{X}'$ and add it to the original latent representation:
\begin{equation}
    \mathbf{z_0}^* = \mathbf{z_0} + \Delta \mathbf{z^*}.
\end{equation}

This single-step addition simulates the real-to-fake gap, bypassing expensive denoising. We decode $\mathbf{z_0}^*$ to $x^*$ to create a "simulated fake" image. We replace a portion of the original fake images with these $x^*$ samples during training. This exposes the detector to a broader range of imprints, enhancing generalization to unseen models.

\subsection{Pipeline}
Having established a method to extrapolate diverse training data via our Noise-based Imprint Simulator, we now detail the NIRNet detector, which is specifically engineered to be sensitive to the generalized imprint we simulate. As shown in Figure~\ref{fig:framework}(c), this hybrid-feature discriminator extracts and concatenates three distinct features—noise-based imprint, frequency, and semantic—before feeding them to a final classification head. The entire model is trained end-to-end on our expanded dataset.

\subsubsection{Noise-based Imprint Extractor}
Our core hypothesis is that the latent imprint (defined in ~\cref{subsec:simulator}) manifests as detectable, pixel-level noise artifacts. To capture this, we employ a Noise-based Imprint Extractor, initialized with the weights from~\cite{TruFor}. We fine-tune this extractor with a specialized auxiliary loss ($\mathcal{L}_{\text{aux}}$), computed on a real image $I_r$ and its corresponding simulated fake image $I_f$ (generated via ~\cref{subsec:expansion}). For each $(I_r, I_f)$ pair, we compute two difference vectors: (1) The noise feature difference $\Delta f = f_f - f_r$ from our Extractor, and (2) The latent representation difference $\Delta z = z_f - z_r$ from the VAE Encoder (~\cref{subsec:simulator}).

\begin{figure}[t]
\vspace{-1mm}
    \centering
    \includegraphics[width=\columnwidth]{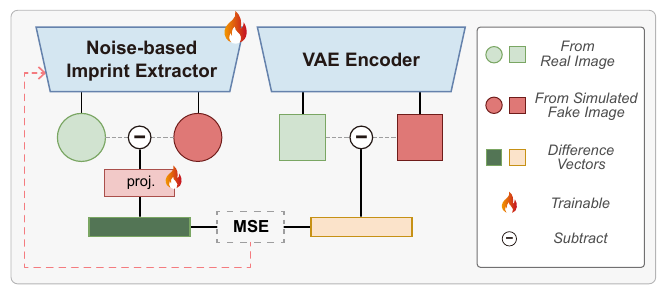}
    \caption{\textbf{Architecture of the Difference-Aware Loss.} This auxiliary loss trains the Noise-based Imprint Extractor to produce noise feature differences that are predictive of the latent representation differences computed by a fixed VAE Encoder. An MLP is trained as a projection module to regress these differences, and the resulting MSE loss guides the extractor to learn pixel-level patterns that correlate with variations in the latent generative space.}
    \label{fig:imprint-extractor}
\vspace{-2mm}
\end{figure}

The loss $\mathcal{L}_{\text{aux}} = \lambda_{\text{diff}} \mathcal{L}_{\text{diff}} + \lambda_{\text{contrast}} \mathcal{L}_{\text{contrast}}$ is twofold:

\begin{itemize}
    \item \textbf{Difference-Aware Loss ($\mathcal{L}_{\text{diff}}$):} As illustrated in Figure~\ref{fig:imprint-extractor}, we train an MLP to predict the latent difference $\Delta z$ using only the noise difference $\Delta f$.
    \begin{equation}
    \mathcal{L}_{\text{diff}} = \text{MSE}(\text{MLP}(\Delta f), \Delta z)
    \end{equation}
    This incentivizes the extractor to learn pixel-level noise features ($f$) that correlate with latent-space changes ($z$).

    \item \textbf{Contrastive Loss ($\mathcal{L}_{\text{contrast}}$):} This is a standard contrastive loss applied to the noise features $f$, which pulls $f_r$ (real) features together while pushing them apart from $f_f$ (fake) features.

\end{itemize}

\subsubsection{Discriminator with Hybrid Features.}
The final discriminator leverages this specialized noise-based imprint feature alongside two other features. A Frequency Extractor from~\cite{AIDE} captures artifacts like excessive smoothness~\cite{AIDE}. A Semantic Extractor (DINOv3 model~\cite{simeoni2025dinov3}) extracts semantic features to identify out-of-distribution characteristics~\cite{UnivFD}.

The entire network is trained end-to-end. The final concatenated features are fed to an MLP discriminator, which is trained with the total loss::
\begin{equation}
    \mathcal{L}_{\text{total}} = \mathcal{L}_{\text{BCE}} + \alpha \cdot \mathcal{L}_{\text{aux}},
\end{equation}
where $\mathcal{L}_{\text{BCE}}$ is the primary binary cross-entropy detection loss and $\alpha$ balances the main task and auxiliary guidance. This hybrid-feature design, guided by $\mathcal{L}_{\text{aux}}$, produces a highly specialized and robust detector.

\section{Experiment}
\label{sec:experiment}

\subsection{Experimental Setup}

\begin{table}[t]
  \centering
  \caption{\textbf{Datasets Overview.} "SD" is Stable Diffusion and "AR" denotes auto-regressive models.}
  \label{table:dataset-overview}
  \vspace{-2mm}
  \captionsetup{width=\linewidth}
  \begin{adjustbox}{width=\linewidth}
  \begin{tabular}{lcccc}
    \toprule
    Dataset & Real / Fake & Source & \# Models & Model Types \\
    \midrule
    GenImage~\cite{genimage} & 48K / 48K & ImageNet & 8 & SD \& GAN \\
    Synthbuster~\cite{synthbuster} & 1K / 9K & RAISE & 9 & SD \& AR \\
    Chameleon~\cite{AIDE} & 14.9K / 11.2K & Internet & unknown & unknown \\
    WildRF~\cite{WildRF} & 500 / 500 & Reddit, FB, X & unknown & unknown \\
    SynthWildx~\cite{cozzolino2024raising} & 500 / 1.5K & X & 3 & SD \& AR \\
    Gen-8K  & 4K / 4K & RAISE, Internet & 2 & SD \& Flow \\
    ForenGen & 0 / 2.2K & Prompt & 4 & AR \& Flow \& Other \\
    \bottomrule
  \end{tabular}
  \end{adjustbox}
  \vspace{-4mm}
\end{table}

\noindent\textbf{Dataset.}
We evaluate our method on public benchmarks and two new datasets (Table~\ref{table:dataset-overview}) to test generalization against diverse and recent generators.
For public benchmarks, we utilize five large-scale and challenging datasets: \textbf{GenImage}~\cite{genimage}, \textbf{Synthbuster}~\cite{synthbuster}, \textbf{Chameleon}~\cite{AIDE}, \textbf{WildRF}~\cite{WildRF}, and \textbf{SynthWildx}~\cite{cozzolino2024raising}. These cover a wide array of generation models (Diffusion, GAN, Auto-Regressive) and realistic scenarios (e.g., social platform images in WildRF). Details are provided in the Supplementary Material.

We introduce two novel test datasets to address recent advancements in image generation:

\begin{itemize}
    \item \textbf{Gen-8K.} This dataset tests recent generators that are not covered by existing benchmarks. It comprises 4,000 real images (RAISE~\cite{RAISE}) and 4,000 generated images (2,000 each) from \textit{FLUX.1-dev}~\cite{flux2024} and \textit{Stable Diffusion 3.5}~\cite{Stable-Diffusion-3-5}.

    \item \textbf{ForenGen.} Designed to evaluate generalization across diverse, unseen T2I generators, ForenGen contains 2,212 images from four generators using aligned prompts from GenEval~\cite{ghosh2023geneval}. These generators are specifically chosen for their unique approaches: (1) \textbf{OmniGen}~\cite{omnigen}: a unified image generation framework that standardizes diverse tasks into an interleaved image–text sequence format; (2) \textbf{Chroma}~\cite{rock2025chroma}: built on a hybrid architecture of multimodal and parallel diffusion transformer blocks through flow matching; (3) \textbf{Harmon}~\cite{Harmon}: a novel unified framework designed for multimodal understanding and generation; (4) \textbf{NOVA}~\cite{NOVA}: a non-quantized auto-regressive model, posing distinct statistical artifacts compared to mainstream diffusion models. The inclusion of these advanced architectures ensures that ForenGen provides a challenging standard for evaluating the generalizability of detectors on unseen generation mechanisms.
\end{itemize}


\begin{table*}[t]
\vspace{-2mm}
\centering
\caption{\textbf{Comparison on GenImage \cite{genimage} and Synthbuster \cite{synthbuster}.} Accuracy (\%) of different detectors (rows) in detecting real and fake images from different generators (columns). The best result and the second-best result are marked in \textbf{bold} and \underline{underline}, respectively.}
\label{table:GenImage-and-Synthbuster}
\begin{adjustbox}{width=\linewidth}
\renewcommand{\arraystretch}{1.25}
  \begin{tabular}{l | cccccccc c | cccccccccc}
    \toprule
    \multirow{2}{*}{\textbf{Method}} & \multicolumn{9}{c|}{\textbf{GenImage}} & \multicolumn{10}{c}{\textbf{Synthbuster}} \\
    \cmidrule(lr){2-10} \cmidrule(lr){11-20}
    & MJ & SDv1.4 & SDv1.5 & ADM & GLIDE & Wukong & VQDM & BigGAN & \textbf{Avg.} & SDv2 & SDv1.3 & GLIDE & Firefly & DALL·E 3 & DALL·E 2 & MJ & SDv1.4 & SDXL & \textbf{Avg.} \\
    \midrule
    CNNSpot~\cite{CNNSpot} & 54.98 & 86.83 & 86.90 & 48.85 & 48.48 & 86.12 & 52.78 & 49.28 & 64.28 & 58.02 & 66.55 & 34.87 & 44.85 & 44.23 & 40.89 & 40.03 & 67.00 & 43.83 & 48.92 \\
    FreDect~\cite{FreDect} & 56.33 & 93.56 & 93.43 & 48.46 & 49.07 & 92.20 & 51.18 & 48.01 & 66.53 & 58.00 & 82.40 & 49.85 & 49.50 & 50.90 & 50.60 & 50.95 & 82.75 & 54.15 & 58.79 \\
    Fusing~\cite{Fusing} & 70.15 & 97.95 & 97.84 & 54.33 & 58.37 & 97.90 & 50.35 & 49.18 & 72.01 & 58.44 & \underline{99.00} & 57.89 & 50.25 & 74.28 & 52.10 & 59.24 & 99.15 & 61.53 & 67.99 \\
    LNP~\cite{LNP} & 55.78 & 97.04 & 96.51 & 48.36 & 48.95 & 94.69 & 51.25 & 48.05 & 67.58 & 64.35 & 86.60 & 49.25 & \underline{54.85} & 56.50 & 52.55 & 50.55 & 85.40 & 55.56 & 61.73 \\
    Lgrad~\cite{LGrad} & 65.88 & 99.62 & 99.38 & 52.63 & 52.24 & 97.98 & 52.68 & 49.80 & 71.27 & 59.79 & 98.05 & 54.04 & 50.80 & 60.64 & 51.29 & 54.69 & 97.60 & 61.04 & 65.33 \\
    DIRE~\cite{DIRE} & 66.48 & 99.63 & 99.43 & 51.38 & 53.78 & 98.87 & 51.70 & 49.73 & 71.38 & 60.84 & 98.25 & 53.25 & 50.49 & 51.19 & 51.40 & 53.45 & 98.20 & 60.04 & 64.12 \\
    UnivFD~\cite{UnivFD} & 62.13 & 99.74 & 99.65 & 50.38 & 56.47 & 98.52 & 92.83 & 50.02 & 76.22 & 56.20 & \textbf{99.70} & 62.15 & 50.65 & 54.90 & 52.70 & 58.30 & \underline{99.45} & 62.55 & 66.29 \\
    NPR~\cite{NPR} & 66.38 & 98.91 & 98.84 & 55.60 & 67.68 & 97.58 & 60.25 & 54.20 & 74.93 & 51.10 & 95.35 & 62.23 & 49.30 & 51.39 & 51.35 & 50.90 & 95.35 & 54.04 & 62.33 \\
    AIDE~\cite{AIDE} & 83.08 & \textbf{99.88} & \underline{99.83} & 73.33 & 94.17 & \textbf{99.84} & 93.08 & 52.24 & 86.93 & 64.70 & 92.90 & 78.60 & 41.15 & 55.95 & 47.20 & 76.05 & 93.90 & 62.10 & 68.06 \\
    DRCT~\cite{chen2024drct} & 91.50 & 94.41 & 95.01 & 79.42 & 89.18 & 94.67 & 90.03 & 81.67 & 89.49 & 69.92 & 97.10 & 51.29 & 51.58 & 66.10 & 48.75 & 48.69 & 95.78 & 44.25 & 63.72 \\
    SAFE~\cite{SAFE} & \underline{95.27} & 99.41 & 99.27 & \underline{82.05} & \underline{96.29} & 98.20 & \underline{96.29} & \textbf{97.84} & \underline{95.58} & \underline{82.15} & 96.55 & \underline{79.35} & 51.40 & \underline{78.05} & \underline{81.10} & \underline{79.85} & 97.25 & \underline{82.60} & \underline{80.92} \\
    \midrule
    \rowcolor{mygray}
    \textit{\textbf{NIRNet (Ours)}} & \textbf{98.18} & \underline{99.84} & \textbf{99.85} & \textbf{99.48} & \textbf{99.77} & \underline{99.78} & \textbf{99.75} & \underline{94.83} & \textbf{98.94} & \textbf{84.30} & 98.85 & \textbf{91.25} & \textbf{75.19} & \textbf{84.80} & \textbf{86.85} & \textbf{91.45} & \textbf{99.48} & \textbf{91.85} & \textbf{89.34} \\
    \bottomrule
  \end{tabular}
\end{adjustbox}
\end{table*}

\begin{table*}[t]
    \centering
    \vspace{1ex}
    \begin{minipage}[t]{0.689\textwidth} 
        \centering
        \captionof{table}{\textbf{Comparison on Chameleon~\cite{AIDE}, SynthWildx~\cite{cozzolino2024raising} and WildRF~\cite{WildRF}.}}
        \label{tab:wild-datasets}
        \vspace{-5pt}
        \footnotesize 
        \renewcommand{\arraystretch}{1.2}
        \begin{adjustbox}{width=\linewidth}
            \begin{tabular}{l c | c c c c | c c c c}
                \toprule
                \multirow{2}{*}{\textbf{Method}} & \multirow{2}{*}{\textbf{Chameleon}} & \multicolumn{4}{c|}{\textbf{SynthWildx}} & \multicolumn{4}{c}{\textbf{WildRF}} \\
                \cmidrule(lr){3-6} \cmidrule(lr){7-10}
                & & DALL·E 3 & Firefly & Midjourney & \textbf{Avg.} & Facebook & Reddit & Twitter & \textbf{Avg.} \\
                \midrule
                CNNSpot~\cite{CNNSpot} & 58.56 & 51.07 & 48.57 & 44.37 & 48.00 & 50.00 & 59.87 & 49.48 & 53.12 \\
                FreDect~\cite{FreDect} & 59.03 & 50.87 & 51.11 & 47.92 & 49.97 & 49.68 & 59.06 & 51.09 & 53.28 \\
                Fusing~\cite{Fusing} & \underline{68.97} & 72.87 & 57.01 & \underline{52.17} & \underline{60.68} & \underline{60.94} & 67.40 & 60.76 & \underline{63.03} \\
                LNP~\cite{LNP} & 61.92 & 54.55 & 52.34 & 46.91 & 51.27 & 52.81 & 60.26 & 53.14 & 55.40 \\
                Lgrad~\cite{LGrad} & 65.82 & 67.24 & \underline{58.13} & 50.45 & 58.61 & 58.13 & 67.87 & \underline{60.91} & 62.30 \\
                DIRE~\cite{DIRE} & 61.75 & 60.79 & 54.27 & 48.63 & 54.56 & 57.18 & 64.33 & 56.95 & 59.49 \\
                UnivFD~\cite{UnivFD} & 58.94 & 63.66 & 50.91 & 49.95 & 54.84 & 57.81 & 66.46 & 58.42 & 60.90 \\
                NPR~\cite{NPR} & 55.81 & 55.06 & 50.50 & 50.45 & 52.00 & 53.13 & 64.13 & 55.05 & 57.44 \\
                AIDE~\cite{AIDE} & 62.99 & \underline{73.49} & 49.79 & 51.36 & 58.21 & 52.50 & 65.87 & 58.71 & 59.03 \\
                DRCT~\cite{chen2024drct} & 65.24 & 52.00 & 52.74 & 50.56 & 51.77 & 48.75 & 61.93 & 56.22 & 55.63 \\
                SAFE~\cite{SAFE} & 59.13 & 49.94 & 49.89 & 49.44 & 49.76 & 50.93 & \underline{74.06} & 52.12 & 59.04 \\
                \midrule
                \rowcolor{mygray}
                \textit{\textbf{NIRNet (Ours)}} & \textbf{81.88} & \textbf{93.14} & \textbf{73.17} & \textbf{93.31} & \textbf{86.54} & \textbf{81.56} & \textbf{85.00} & \textbf{89.01} & \textbf{85.19} \\
                \bottomrule
            \end{tabular}
        \end{adjustbox}
    \end{minipage}
    \hfill
    \begin{minipage}[t]{0.281\textwidth}
        \centering
        \captionof{table}{\textbf{Comparison on Gen-8K.}}
        \label{tab:gen8k}
        \vspace{-5pt}
        \footnotesize
        \renewcommand{\arraystretch}{1.25}
        \begin{adjustbox}{width=\linewidth}
        \begin{tabular}{l cc c}
            \toprule
            \textbf{Method} & SD 3.5 & FLUX & \textbf{Avg.} \\
            \midrule
            CNNSpot~\cite{CNNSpot} & 55.93 & 43.28 & 49.60 \\
            FreDect~\cite{FreDect} & 55.75 & 51.30 & 53.53 \\
            Fusing~\cite{Fusing} & 59.35 & 58.23 & 58.79 \\
            LNP~\cite{LNP} & 59.08 & 55.53 & 57.30 \\
            Lgrad~\cite{LGrad} & 55.45 & 54.98 & 55.21 \\
            DIRE~\cite{DIRE} & 57.38 & 54.78 & 56.08 \\
            UnivFD~\cite{UnivFD} & 57.83 & 57.50 & 57.66 \\
            NPR~\cite{NPR} & 59.30 & 60.15 & 59.73 \\
            AIDE~\cite{AIDE} & 56.65 & \underline{63.00} & 59.83 \\
            DRCT~\cite{chen2024drct} & \underline{66.48} & 60.33 & \underline{63.40} \\
            SAFE~\cite{SAFE} & 52.80 & 50.10 & 51.45 \\
            \midrule
            \rowcolor{mygray}
            \textit{\textbf{NIRNet (Ours)}} & \textbf{93.13} & \textbf{94.95} & \textbf{94.04} \\
            \bottomrule
        \end{tabular}
        \end{adjustbox}
    \end{minipage}
    \vspace{-2mm}
\end{table*}

\noindent\textbf{Implementation details.}
For the Noise-based Imprint Simulator, we utilize the VAE from the LDM~\cite{LDM}. We fit the noise distribution using 2,000 images from RAISE~\cite{RAISE}, reconstructing them via pre-trained diffusion models (SD 2.1, SD-Turbo, SDXL, SDXL-Turbo) configured with 400 inference steps, strength 0.1, and guidance 0.0. The Noise-based Imprint Extractor is initialized using Noiseprint++~\cite{TruFor} and trained with $\lambda_{\text{diff}}=0.2$, $\lambda_{\text{contrast}}=1.0$ and $\alpha=0.2$ .

Following the generalization setting in~\cite{AIDE}, all models are trained on 162,000 real and 162,000 fake images (Stable Diffusion v1.4 from GenImage). For NIRNet data expansion, we sample 2\% of real images and synthesize 5 variants each via the Noise-based Imprint Simulator, replacing 16,200 original fake images. This maintains the training set size of 162,000 real and 162,000 fake images (145,800 original and 16,200 synthetic).

Augmentations follow \cite{CNNSpot}, including random JPEG compression (quality factor QF $\sim \text{Uniform}(30, 100)$) and Gaussian blur ($\sigma \sim \text{Uniform}(0.1, 3.0)$). Training uses AdamW~\cite{AdamW} optimizer with a learning rate of $1 \times 10^{-4}$. Experiments run on 8 NVIDIA RTX 4090 GPUs with a batch size of 32 for 5 epochs.

\noindent\textbf{Evaluation Metric.} We report classification accuracy (Acc) following \cite{CNNSpot,AIDE}. Unless otherwise specified, all results are averaged across both real and fake images. For the ForenGen dataset, which only contains fake images, we report the accuracy of correctly identifying them as fake (i.e., the True Positive Rate).

\subsection{State-of-the-art Comparison} 
To ensure a fair comparison, we only consider the methods with publicly available code for evaluation. We compare with 11 methods including
CNNSpot~\cite{CNNSpot}, FreDect~\cite{FreDect}, Fusing~\cite{Fusing}, LNP~\cite{LNP}, LGrad~\cite{LGrad}, DIRE~\cite{DIRE}, UnivFD~\cite{UnivFD}, NPR~\cite{NPR}, AIDE~\cite{AIDE}, DRCT~\cite{chen2024drct} and SAFE~\cite{SAFE}. A brief description of these methods can be found in the supplementary material.
All methods are trained on GenImage/SD v1.4. The data expansion used in NIRNet also leads to a consistent training set size across methods.

\noindent\textbf{Result on the GenImage and Synthbuster datasets.} Table~\ref{table:GenImage-and-Synthbuster} shows that many detectors overfit to the training architecture (SD v1.4), with large drops on dissimilar models like ADM and BigGAN (GAN-based). This suggests they learn superficial artifacts rather than a universal one. In contrast, NIRNet achieves the highest average accuracy on both GenImage (98.94\%) and Synthbuster (89.34\%). Its outstanding performance on distinct models, such as GLIDE (91.25\%) and the newer SDXL (91.85\%), suggests our noise-based imprint is a more fundamental and transferable cue.

\begin{table}[t]
\vspace{-1mm}
\caption{\textbf{Comparison on ForenGen.} Since this dataset only contains fake images, accuracy refers to the rate of correctly classifying them as fake (True Positive Rate).}
\label{tab:forengen}
\vspace{-2mm}
    \centering
    \vspace{1ex}
    \footnotesize
    \renewcommand{\arraystretch}{1.25}
    \begin{adjustbox}{width=0.9\columnwidth}
    \begin{tabular}{l cccc c}
        \toprule
        \textbf{Method} & Chroma & NOVA & Harmon & OmniGen & \textbf{Avg.} \\
        \midrule
        CNNSpot~\cite{CNNSpot} & 4.88 & 15.01 & 15.76 & 23.69 & 14.84 \\
        FreDect~\cite{FreDect} & 3.25 & 10.67 & 10.51 & 15.37 & 9.95 \\
        Fusing~\cite{Fusing} & 34.53 & 55.69 & 79.52 & 48.82 & 54.64 \\
        LNP~\cite{LNP} & 3.98 & 16.64 & 18.12 & 15.55 & 13.57 \\
        Lgrad~\cite{LGrad} & 14.47 & 34.18 & 63.59 & 30.19 & 35.61 \\
        DIRE~\cite{DIRE} & 12.48 & 33.28 & 57.43 & 46.11 & 37.33 \\
        UnivFD~\cite{UnivFD} & 13.02 & 32.55 & \textbf{96.92} & 60.75 & 50.81 \\
        NPR~\cite{NPR} & 28.21 & \underline{79.92} & 75.54 & 83.01 & 66.67 \\
        AIDE~\cite{AIDE} & 13.20 & 34.18 & 21.19 & 73.59 & 35.54 \\
        DRCT~\cite{chen2024drct} & \underline{56.25} & 70.83 & 64.29 & 85.42 & \underline{69.20} \\
        SAFE~\cite{SAFE} & 10.82 & 9.37 & 36.54 & \textbf{99.45} & 39.05 \\
        \midrule
        \rowcolor{mygray}
        \textit{\textbf{NIRNet (Ours)}} & \textbf{56.60} & \textbf{96.21} & \underline{89.49} & \underline{94.58} & \textbf{84.22} \\
        \bottomrule
    \end{tabular}
    \end{adjustbox}
\vspace{-4mm}
\end{table}

\noindent\textbf{Result on In-the-Wild Datasets.} To assess real-world applicability, we evaluated NIRNet on three challenging "in-the-wild" datasets: Chameleon, SynthWildx, and WildRF, with results compiled in Table~\ref{tab:wild-datasets}. These benchmarks test a detector's resilience to unknown generators and widespread post-processing artifacts common on social media. Notably, methods like SAFE, which performed competitively on curated datasets, fail significantly here, with accuracy dropping to near chance on SynthWildx (49.76\%). This highlights a critical gap between performance on clean and real-world data. In contrast, NIRNet achieves robust SOTA accuracy, outperforming the second-best methods by 12.91\% (Chameleon), 25.86\% (SynthWildx), and 22.16\% (WildRF). While NIRNet's performance is dominant, the overall accuracy is lower than on curated benchmarks, which is likely due to heavy "in-the-wild" compression and artifacts obscuring noise-based imprint, challenging all detectors.

\noindent\textbf{Result on Novel and Unseen Architectures.} We report the performance on Gen-8K and ForenGen datasets in Table~\ref{tab:gen8k} and~\ref{tab:forengen}. These datasets are specifically designed to test generalization against the latest and most diverse generator architectures. On Gen-8K, which includes recent models such as SD 3.5 and FLUX.1-dev, NIRNet reaches over 93\% accuracy, while most prior methods fall below 65\%. This demonstrates strong adaptation to next-generation generators without prior exposure. On the architecturally diverse ForenGen dataset, NIRNet achieves 84.22\% average accuracy, surpassing all baselines and showing robustness to non-diffusion and autoregressive models by capturing fundamental imprint rather than model-specific artifacts.

\subsection{Ablation Study}

\subsubsection{Analysis of Core Component Contributions}

To assess the contributions of the Noise-based Imprint Simulator (NIS) and the Noise-based Imprint Extractor (NIE), we conducted an ablation study on the GenImage dataset~\cite{genimage} across six configurations: (1) \textbf{Base}: frequency and semantic features only, (2) \textbf{Base + NIE}: The Base model with the NIE module. (3) \textbf{Base + NIS}: The Base model with the NIS module. (4) \textbf{Full model (Base + NIS + NIE)}: The complete NIRNet framework, (5) \textbf{NIE only}: The NIE module without the base features, and (6) \textbf{NIS + NIE}: The NIS and NIE modules combined, also without the base features. Table~\ref{table:ablation} summarizes the results for each configuration on the dataset GenImage~\cite{genimage}.

The Base model achieved a respectable accuracy of 87.85\%. This performance serves as a reference point for our innovations. Integrating the NIE module increased accuracy to 91.24\%, underscoring the rich discriminative information contained within pixel-level noise patterns, which are often overlooked by conventional methods. Even more remarkably, when the Base model was augmented with our data expansion strategy via the NIS module, performance further improved to 95.57\%. This gain supports our hypothesis: simulating imprint from a fused distribution effectively reduces overfitting and enhances its generalization to unseen generative artifacts.

\begin{table}[tb!]
    \vspace{-1mm}
    \centering
    \caption{\textbf{Ablation studies on Noise-based Imprint Simulator (NIS) and Noise-based Imprint Extractor (NIE) modules.} Base indicates incorporating only frequency and semantic features.}
    \label{table:ablation}
    \vspace{-1mm}
    \footnotesize
    \captionsetup{width=\linewidth}  
    \begin{tabular}{ccc|c}
    \toprule
    \multicolumn{3}{c|}{\textbf{Module}}                         & \multirow{2}{*}{\textbf{Mean}} \\ 
    \textbf{Base}        & \textbf{NIS}        & \textbf{NIE}        &                                   \\ \midrule
    \textbf{\ding{55}} & \textbf{\ding{55}} & \textbf{\ding{51}}          & 83.62                        \\
    \textbf{\ding{55}} & \textbf{\ding{51}} & \textbf{\ding{51}}          & 88.51                        \\ \midrule
    \textbf{\ding{51}} & \textbf{\ding{55}}          & \textbf{\ding{55}}          & 87.85                        \\
    \textbf{\ding{51}} & \textbf{\ding{55}} & \textbf{\ding{51}}          & 91.24                        \\
    \textbf{\ding{51}} & \textbf{\ding{51}} & \textbf{\ding{55}} & 95.57                        \\ \midrule
    \rowcolor{mygray}
    \textbf{\ding{51}} & \textbf{\ding{51}} & \textbf{\ding{51}} & \textbf{98.94}                    \\ \bottomrule
    \end{tabular}
\end{table}

\begin{table}[tb!]
    \centering
    \caption{\textbf{Ablation studies on the loss functions of the  Noise-based Imprint Extractor (NIE) module.} We evaluate the impact of $L_{\text{diff}}$ and $L_{\text{contrast}}$ on the GenImage dataset.}
    \label{table:ablation_loss}
    \vspace{-1mm}
    \footnotesize
    \captionsetup{width=\linewidth}
    \begin{adjustbox}{width=0.8\linewidth}
    \begin{tabular}{cc|c}
    \toprule
    \multicolumn{2}{c|}{\textbf{NIE Loss Component}} & \multirow{2}{*}{\textbf{Mean}} \\ 
    $L_{\text{diff}}$ ($\lambda_{\text{diff}}=0.2$) & $L_{\text{contrast}}$ ($\lambda_{\text{contrast}}=1.0$) & \\ 
    \midrule
    \textbf{\ding{55}} & \textbf{\ding{55}} & 95.62 \\ 
    \textbf{\ding{55}} & \textbf{\ding{51}} & 96.85 \\ 
    \textbf{\ding{51}} & \textbf{\ding{55}} & 97.29 \\
    \midrule
    \rowcolor{mygray}
    \textbf{\ding{51}} & \textbf{\ding{51}} & \textbf{98.94} \\ 
    \bottomrule
    \end{tabular}
    \end{adjustbox}
    \vspace{-3mm}
\end{table}

The synergy between our components is evident in the full NIRNet framework, which achieved an accuracy of 98.94\%, an 11\% improvement over the Base, demonstrating that our modules are not just additive but complementary. To further isolate the power of our noise-based approach, we evaluated the NIE and NIS modules without the base features. The combination of NIS and NIE alone reached an accuracy of 88.51\%, outperforming the Base model (87.85\% vs. 88.51\%). This finding is particularly significant, as our compact noise-focused modules (\~136M) prove more effective than the Base Model (\~1420M) that relies on significantly larger, pre-trained networks for semantic and frequency analysis. These results strongly indicate that both the NIS and NIE are pivotal to NIRNet's success, forming a powerful and efficient detection paradigm centered on the fundamental concept of noise-based imprint.

\subsubsection{Analysis of Noise-based Imprint Extractor Loss}
We validated the Noise-based Imprint Extractor's (NIE) objective functions, $L_{\text{diff}}$ and $L_{\text{contrast}}$, in Table~\ref{table:ablation_loss}. As a baseline, we trained the full architecture (Base + NIS + NIE) with both $\lambda_{\text{diff}}$ and $\lambda_{\text{contrast}}$ set to zero, yielding 95.62\% accuracy. This is only a marginal improvement (0.05\%) over the Base + NIS model (95.57\%; Table~\ref{table:ablation}), which excludes NIE. This confirms that NIE’s effectiveness arises not from its architecture but from its specialized loss functions.

When applied individually, both losses significantly improved performance: $L_{\text{contrast}}$ achieved 96.85\%, and $L_{\text{diff}}$ reached 97.29\%. Using both jointly produced the best result of \textbf{98.94\%}, demonstrating their complementary roles in extracting discriminative noise imprint. The selected $\lambda$ values are provided in the Supplementary Material.


\begin{table}[tb!]
\vspace{-1mm}
    \centering
    \caption{\textbf{Ablation studies on the impact of model diversity in Noise-based Imprint Simulator.} The table shows the model performance (accuracy \%) on the GenImage and ForenGen datasets when using various combinations of Stable Diffusion models.}
    \label{table:model_ablation}
    \vspace{-1mm}
    \footnotesize
    \captionsetup{width=\linewidth}
    \begin{adjustbox}{width=1.0\linewidth}
    \begin{tabular}{@{} ccccc | cc @{}}
    \toprule
    \multicolumn{5}{c|}{\textbf{Model}} & \multirow{2}{*}{\textbf{GenImage}} & \multirow{2}{*}{\textbf{ForenGen}} \\
    \textbf{SD-1-4} & \textbf{SDXL} & \textbf{SD-2-1} & \textbf{SD-Turbo} & \textbf{SDXL-Turbo} & & \\
    \midrule
    \ding{55} & \ding{55} & \ding{55} & \ding{55} & \ding{55} & 91.24 & 55.76 \\
    \midrule
    \ding{51} & \ding{55} & \ding{55} & \ding{55} & \ding{55} & 94.15 & 60.38 \\
    \midrule
    \ding{55} & \ding{51} & \ding{55} & \ding{55} & \ding{55} & 92.89 & 62.43 \\
    \ding{55} & \ding{55} & \ding{51} & \ding{55} & \ding{55} & 93.21 & 68.52 \\
    \ding{55} & \ding{55} & \ding{55} & \ding{51} & \ding{55} & 93.13 & 59.96 \\
    \ding{55} & \ding{55} & \ding{55} & \ding{55} & \ding{51} & 94.92 & 65.88 \\
    \midrule
    \ding{55} & \ding{51} & \ding{51} & \ding{55} & \ding{55} & 93.45 & 72.94 \\
    \ding{55} & \ding{55} & \ding{55} & \ding{51} & \ding{51} & 96.28 & 76.85 \\
    \midrule
    \rowcolor{mygray}
    \ding{55} & \ding{51} & \ding{51} & \ding{51} & \ding{51} & \textbf{98.94} & \textbf{84.22} \\
    \bottomrule
    \end{tabular}
    \end{adjustbox}
    \vspace{-3mm}
\end{table}

\subsubsection{Impact of Model Diversity in Noise-based Imprint Simulator}

We next evaluated how model diversity in the Noise-based Imprint Simulator affects generalization, following our hypothesis that fused imprint distributions better simulate unseen artifacts. Results on GenImage and the challenging ForenGen benchmark are shown in Table~\ref{table:model_ablation}.

Without Noise-based Imprint Simulator, accuracy drops to 91.24\% on GenImage and 55.76\% on ForenGen. Introducing imprint from even a single model markedly improves performance. More importantly, the results reveal a clear monotonic trend: increasing the diversity of contributing models consistently strengthens generalization. For instance, two models elevate ForenGen accuracy above 72\%, while using all four models yields the best performance—98.94\% on GenImage and 84.22\% on ForenGen. Furthermore, to ensure a strict comparison, we added an experiment using only SD v1.4 , which achieved 94.15\% (GenImage) and 60.38\% (ForenGen). This confirms the method's effectiveness even in this constrained setting.

These findings validate our hypothesis: aggregating imprint from diverse generators enables Noise-based Imprint Simulator to form a richer statistical representation, mitigating overfitting to specific artifacts and substantially improving generalization.

\begin{figure}[t]
\vspace{-1mm}
    \centering
    \includegraphics[width=1.0\columnwidth]{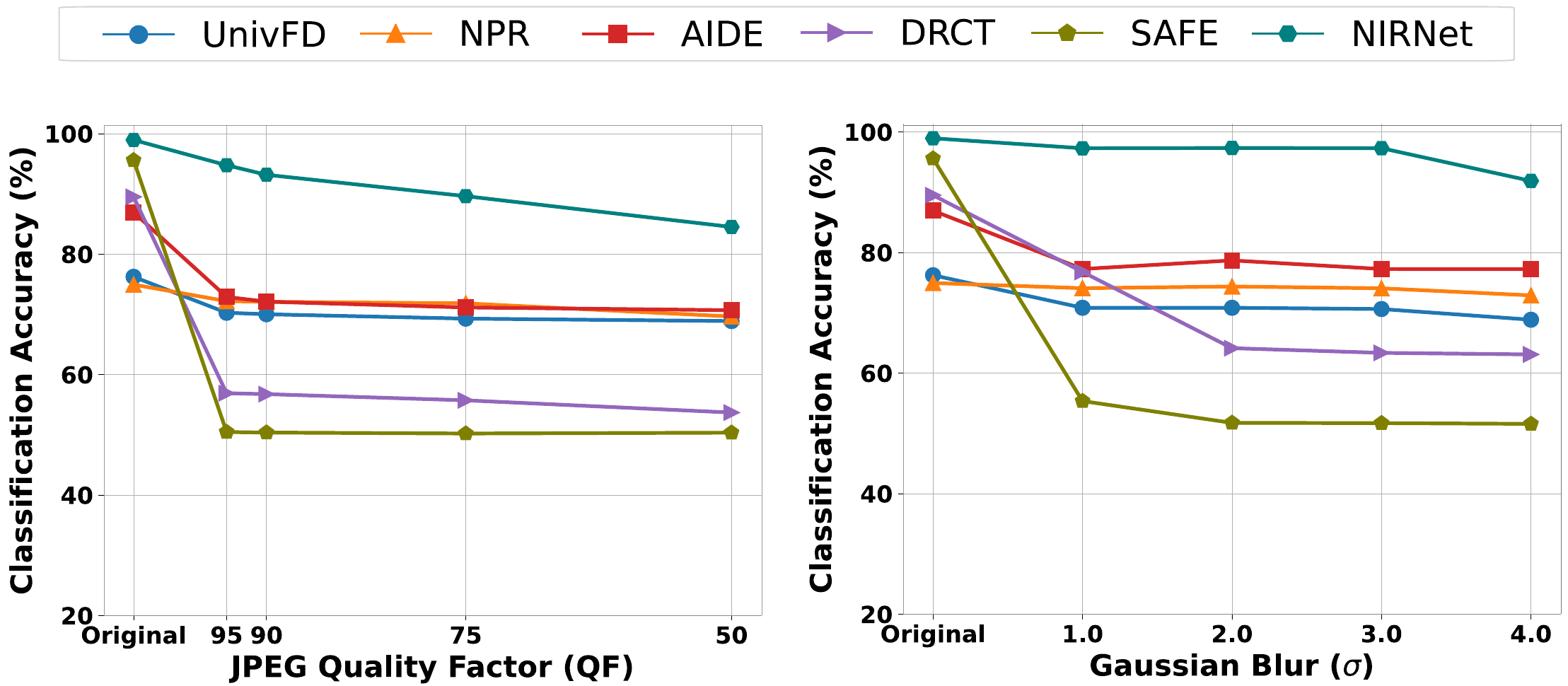}
    \caption{\textbf{Robustness analysis on GenImage}}
    \label{fig:robustness}
\vspace{-3mm}
\end{figure}

\subsection{Robustness to Perturbations}

We further assessed robustness to common real-world perturbations, including JPEG compression (quality 95, 90, 75, 50) and Gaussian blur ($\sigma$=1.0, 2.0, 3.0, 4.0), following the protocol of~\cite{CNNSpot}. Results are shown in Figure~\ref{fig:robustness}, with detailed values in the Supplementary Material.

Across all perturbation levels, NIRNet consistently outperforms prior methods. Under JPEG quality 75, NIRNet maintains 89.61\% accuracy, far exceeding NPR’s 71.81\%. Under Gaussian blur with $\sigma$=3.0, NIRNet achieves 97.30\%, compared to AIDE’s 77.25\%. This robustness stems from two factors: (1) Noise-based Imprint Simulator captures intrinsic noise patterns across diverse generators, reducing sensitivity to superficial distortions introduced by compression or blur; and (2) Noise-based Imprint Extractor focuses on fundamental low-level differences between camera-captured and AI-generated content, which remain stable under post-processing. These results highlight NIRNet’s practicality and resilience in real-world scenarios.

\section{Conclusion}
\label{sec:conclusion}
We introduced NIRNet, a novel and generalizable framework for detecting AI-generated images, leveraging inherent noise-based imprint from generative models. Our approach uses a Noise-based Imprint Simulator to expand training data and a hybrid detection pipeline integrating noise, frequency, and semantic features. NIRNet achieved state-of-the-art performance across seven benchmarks, demonstrating strong cross-model generalization, confirming that noise imprints are a powerful and resilient cue for distinguishing real from synthetic images.

\vspace{-5mm}
\paragraph{Limitations and Future Work.} Future work could enhance the universality of our Noise-based Imprint Simulator by incorporating imprint from diverse generative paradigms, such as GANs and AR models. Furthermore, investigating NIRNet's resilience against adaptive adversarial attacks, potentially through countermeasures like adversarial training or certified defenses, remains a critical future direction.

{
    \small
    \bibliographystyle{ieeenat_fullname}
    \bibliography{main}
}
\end{document}